\documentclass[11pt]{article}
\usepackage{eamt26}
\usepackage{times}
\usepackage{url}
\usepackage{latexsym}
\usepackage[small,bf]{caption} 
\usepackage[T1]{fontenc}
\DeclareUnicodeCharacter{2581}
{\textunderscore}
\DeclareUnicodeCharacter{2260}{\ensuremath{\neq}}
\usepackage{listings}
\usepackage{xcolor}
    
\title{The Impact of Vocabulary Overlaps on Knowledge Transfer in Multilingual Machine Translation}

\author{
Oona Itkonen \and Jörg Tiedemann \\
University of Helsinki \\
}

\date{}

\begin{document}
\maketitle
\begin{abstract}
  Knowledge transfer, especially across related languages, has been found beneficial for multilingual neural machine translation (MNMT), but some aspects are still under-explored and deserve further investigation. A joint vocabulary is most often applied to form a uniform word embedding space, but since the impact of a disjoint vocabulary on model performance is far less studied, there is no consensus on how much knowledge transfer is mainly due to vocabulary overlap. In this paper, we present systematic experiments with joint and disjoint vocabularies, and auxiliary languages related and unrelated to the source language. We design this experiment in an out-of-domain setup in order to emphasize transfer and the impact of the auxiliary language. As expected, we yield better results with more extensive vocabulary overlaps typical for related languages, but our experiments also show that domain-match and language relatedness are more important than a joint vocabulary. 
\end{abstract}

\section{Introduction}

Neural machine translation models can take advantage of transferring knowledge between languages in a multilingual setting, but many factors contribute to what extent the transfer is enabled and what kind of information is shared between the languages. Tokenizer configuration, vocabulary settings and other details of the pipeline substantially affect the performance of a model, and therefore arguably have an influence on the knowledge transfer within the model. However, factors not dependent on the training pipeline, such as relatedness of the languages, can be equally, if not more significant. 

\begin{figure}[t]
  \includegraphics[width=\columnwidth]{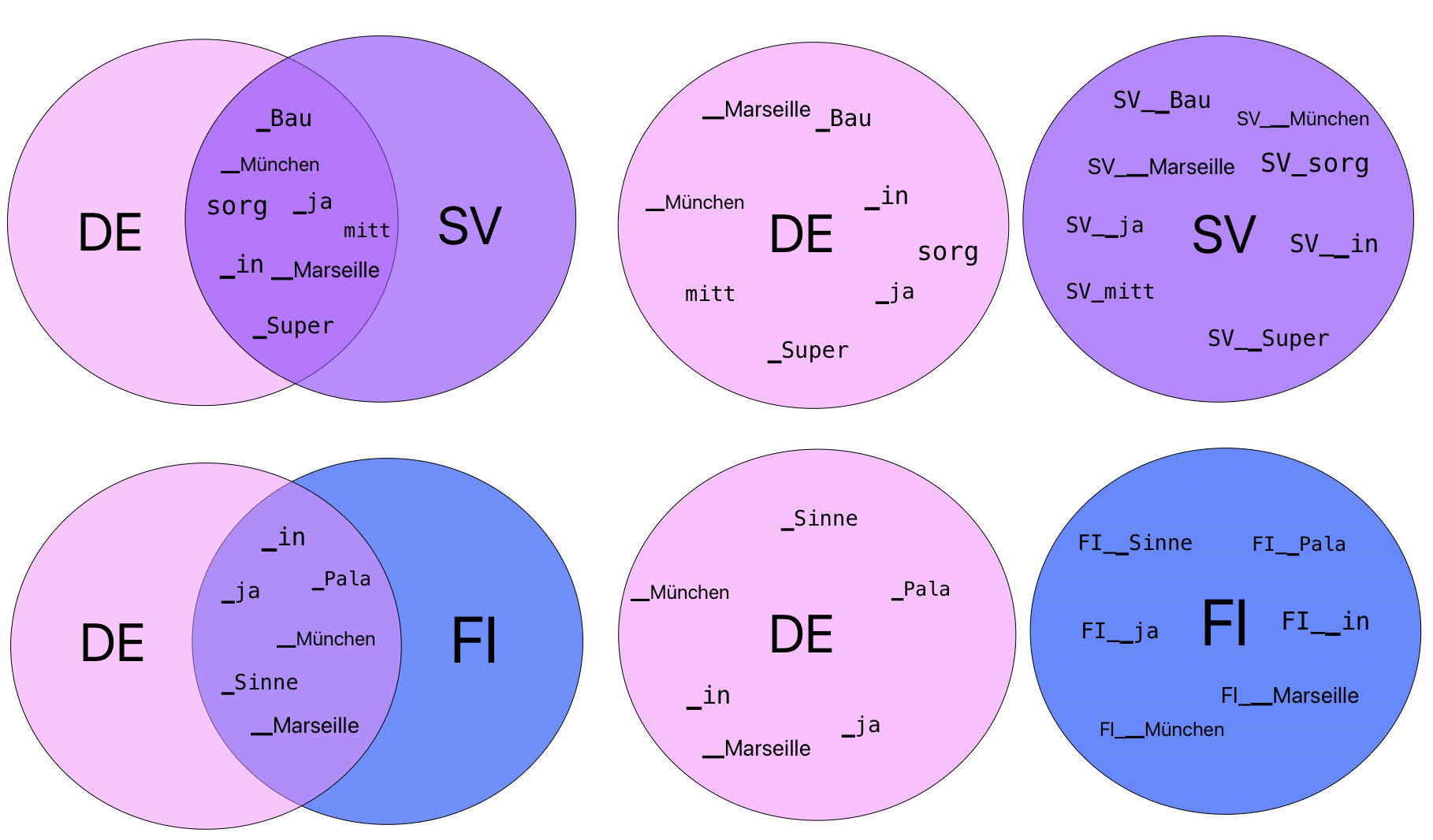}
  \caption{Joint and disjoint vocabularies.}
  \label{fig:overlapofoverlaps}
\end{figure}

Even though knowledge transfer has been found to bring benefit to MNMT, the technicalities of the phenomenon are not completely understood ~\cite{stap-etal-2023-viewing}. Related languages are often grouped together for better results, but the majority of research is done in a knowledge distillation setup ~\cite{zoph-etal-2016-transfer,dabre-etal-2017-empirical,aji-etal-2020-neural}. We, however, approach this issue by training multilingual models where we introduce in-domain data through an auxiliary language to find out how much it can benefit the translation task, and therefore how much of it is transferred to another language pair that is trained on out-of-domain data. We experiment with two auxiliary source languages, Swedish and Finnish, and compare their impact on translations from German to English.

Our main interest is to find out how much of the knowledge transfer in our experiments is in fact lexical. Sharing a vocabulary is nowadays the general approach to multilinguality in any Natural Language Processing tasks, including Machine Translation ~\cite{10.1145/3406095,deshpande-etal-2022-bert}. A joint vocabulary has been found beneficial ~\cite{sennrich-etal-2016-neural,10.1145/3406095}, as it can promote the uniformity of the embedding space ~\cite{wu-monz-2023-beyond}. This is due to the vocabulary overlap that emerges naturally when a multilingual subword tokenizer is applied in the pipeline. 

Studies experimenting with a disjoint vocabulary are rather sparse and there is not much research dedicated to the comparison of different vocabulary settings in the field of Machine Translation. The benefit of vocabulary overlap is intuitive to related languages that share a lot of lexical items, but research within multilingual language models ~\cite{kallini-etal-2025-false} suggests, that the impact of a joint vocabulary might not be as straightforward for unrelated languages. In the case of a disjoint vocabulary, however, any benefit from lexical sharing is disabled, but information needs to be transferred between the languages in the shared hidden layers of the network. By experimenting with joint and disjoint vocabularies, we aim to find out how much knowledge transfer is due to vocabulary overlap.

As expected, we yield better results with more extensive vocabulary overlap typical for related languages, though we find a joint vocabulary beneficial for both related and unrelated languages. However, our experiments show that knowledge transfer persists with a disjoint vocabulary and as our models trained with Swedish and German consistently outperform models trained with Finnish and German, we argue that language relatedness is more significant for a successful knowledge transfer than vocabulary overlap. The scripts and configurations to run the experiments presented in this paper are made public on GitHub.\footnote{\url{https://github.com/oonaelektra/BA-thesis}}

\section{Related Work}

\subsection{Multilingual Machine Translation}

The aim of multilingual NMT is to translate between several languages within the same model ~\cite{firat-etal-2016-multi,ha-etal-2016-toward,10.1145/3406095}. There are different takes on what should be shared between languages in a multilingual training pipeline ~\cite{10.1145/3406095,K2020Cross-Lingual,escolano-etal-2021-multilingual}. Possible benefit from knowledge transfer encourages to rather share than keep the languages distinct. Most often, it is the low resource languages that are getting the biggest benefit from a multilingual setting, and more high resource languages might even get compromised results ~\cite{conneau-etal-2020-unsupervised}.

Despite this curse of multilinguality, for some, the utter goal of MNMT might be to build a model capable of any translation task regardless of language or domain. Improving zero-shot translation is therefore an attractive topic to many researchers. A multilingual model can be build to translate from many languages to one, or the opposite, or in a many-to-many setting combining the two. A zero-shot translation, however, refers to a situation, where a model is producing translation for two languages not introduced as a pair to the model, something that is only enabled by multilinguality and knowledge transfer across translation tasks ~\cite{johnson-etal-2017-googles}. Multilinguality can however, bring benefit in less ambitious settings as well, and it is argued that related languages benefit most from each other ~\cite{10.1145/3406095}. Another perspective is therefore, not to aim for a model capable of processing as many languages as possible but rather building models concentrating on which languages are best grouped together. 

\subsection{Knowledge Transfer}
Cross-lingual knowledge transfer refers to the ability of a multilingual model to transfer information between languages rather than treating every language separately. The phenomenon naturally emerges in a multilingual model due to shared hidden layers of the model and vocabulary overlaps. Cross-lingual knowledge transfer has also been studied outside of MNMT, concentrating on multilingual language models and various downstream tasks ~\cite{dufter-schutze-2020-identifying,deshpande-etal-2022-bert}. 

Within MNMT, however, the most dominant perspective of research over the past decade has been improving translation quality for low resource languages by using knowledge distillation methods. Many studies use a parent model trained with a different language pair to transfer knowledge to a child model, which then processes the languages the study was actually targeting ~\cite{zoph-etal-2016-transfer,dabre-etal-2017-empirical,nguyen-chiang-2017-transfer,kocmi-bojar-2018-trivial}. While experimenting with different ways of benefiting from knowledge transfer, its detailed mechanisms as a phenomenon are still not completely understood.~\cite{stap-etal-2023-viewing}.

Transfer between languages can have either a positive or a negative impact to the performance of the model. The negative impact, or so called interference has been studied in MNMT by Shaham et al.~\shortcite{shaham-etal-2023-causes}. They suggest temperature sampling to overcome it. Intuitively, the nature of the transfer, positive or negative, will be reflected in the evaluation scores, but Stap et al.~\shortcite{stap-etal-2023-viewing} argue that BLEU or any other evaluation score is not a reliable indicator of knowledge transfer. They present a new way of studying knowledge transfer, the representational transfer potential (RTP) measuring representational similarities between languages. 

The similarity between languages, however, can be measured on various different levels starting from orthography and script going all the way to morphosyntactic and semantic properties. There is no consensus on the field, which of these are most relevant for knowledge transfer, as there are even results contradicting the observation that relatedness is an important factor ~\cite{kocmi-bojar-2018-trivial}. Dhar and Bisazza~\shortcite{dhar-bisazza-2021-understanding} claim, that syntactic transfer is rather shallow and therefore not important for knowledge transfer. Meyer and Buys~\shortcite{meyer-buys-2024-systematic} argue, that orthography plays a crucial role, however Sannigrahi and Bawden~\shortcite{sannigrahi-bawden-2023-investigating} did not find transliteration significantly important when processing languages of different script. 

Famously, K et al.~\shortcite{K2020Cross-Lingual} state that a joint vocabulary is not necessary for knowledge transfer, which is backed up by Kim et al.~\shortcite{kim-etal-2019-effective}. Despite their findings, the majority of existing research, eg. ~\cite{sennrich-etal-2016-neural,kudo-2018-subword,wu-dredze-2019-beto,chung-etal-2020-improving,patil-etal-2022-overlap,sun-etal-2022-alternative,stap-etal-2023-viewing} and others, do however recommend a joint vocabulary and find lexical overlapping very much beneficial for knowledge transfer.

\subsection{Vocabulary Overlap}
\label{sec:background of vocab overlap}
Within MNMT vocabulary overlap refers to the set of words or subwords shared between two languages. Considering this in the broad sense of lexical sharing, vocabulary overlap is expected to be bigger for related languages and smaller for more distant languages. The both extremes, two languages having the exact same lexical items and two languages having absolutely no shared words or subwords are probably very rare, if not nonexistent, at least for languages that share the same script. Vocabulary overlap can therefore be a purely linguistic question, but we think of it more precisely as the actual overlapping tokens in a vocabulary of a MNMT model. This is sometimes referred to as ``token overlap'', but we stick to vocabulary overlap for clarity. 

Vocabulary overlap itself has not been studied much within MNMT, but a joint vocabulary is most often applied ~\cite{johnson-etal-2017-googles,wu-monz-2023-beyond} and the use of a multilingual subword-tokenizer naturally enables vocabulary overlap. Byte-Pair-Encoding is a widely used subword-tokenizer algorithm, that creates the subwords based on frequency and the vocabulary size. The algorithm itself was created already in the 1990's ~\cite{10.5555/177910.177914}, and it has been used for NLP applications for at least a decade ~\cite{sennrich-etal-2016-neural}, though other algorithms, such as Unigram language models have been proposed and used as well~\cite{kudo-2018-subword}. 

As Kallini et al.~\shortcite{kallini-etal-2025-false} suggest, given the vocabularies $V^1$ and $V^2$ for languages $L^1$ and $L^2$, respectively, the vocabulary overlap of $V^1$ and $V^2$ can be presented as $O = V^1 \cap V^2$. The size of a joint vocabulary $V^{joint}$ for $L^1$ and $L^2$, therefore depends on the size of the vocabulary overlap $O$. More formally, $|V^{joint}| = |V^1| + |V^2| - |O|$. However, this is true for a disjoint vocabulary as well, as $|O| = 0$, when there is no overlap. 

Vocabulary overlap enables joint embeddings between languages and specifically the semantically similar overlapping tokens are expected to bring most advantage to the model. However, the body of vocabulary overlap can consist of any kinds of tokens, semantically similar or not. The latter can include tokens that happen to share the same surface form, but have completely different meaning between the languages, and in an ideal situation should therefore be mapped separately to the embedding space. The overlap can also include subwords, that may not even carry a significant meaning in one or both languages. Kallini et al.~\shortcite{kallini-etal-2025-false} have experimented with only including either semantically similar or dissimilar tokens in the vocabulary overlap and find that the significance of the semantic similarity of the tokens is rather restricted, but that the semantics are more relevant for unrelated languages.

As discussed in the previous section, unrelated languages may not benefit from knowledge transfer as much as closely related and similar languages. In such cases, vocabulary overlap is also expected to be smaller and include less semantically appropriate overlap. Wu and Monz~\shortcite{wu-monz-2023-beyond} have overcome this by merging an equivalence graph to the embedding space, thus managing to pull closer the embeddings of tokens that do share meaning but have completely dissimilar surface forms between languages. As we do not concentrate on the semantic property of the overlap in this study, we also restrict the manipulation of our pipeline to data and tokenizer design, as described below. 

\begin{figure*}
\includegraphics[width=\textwidth]{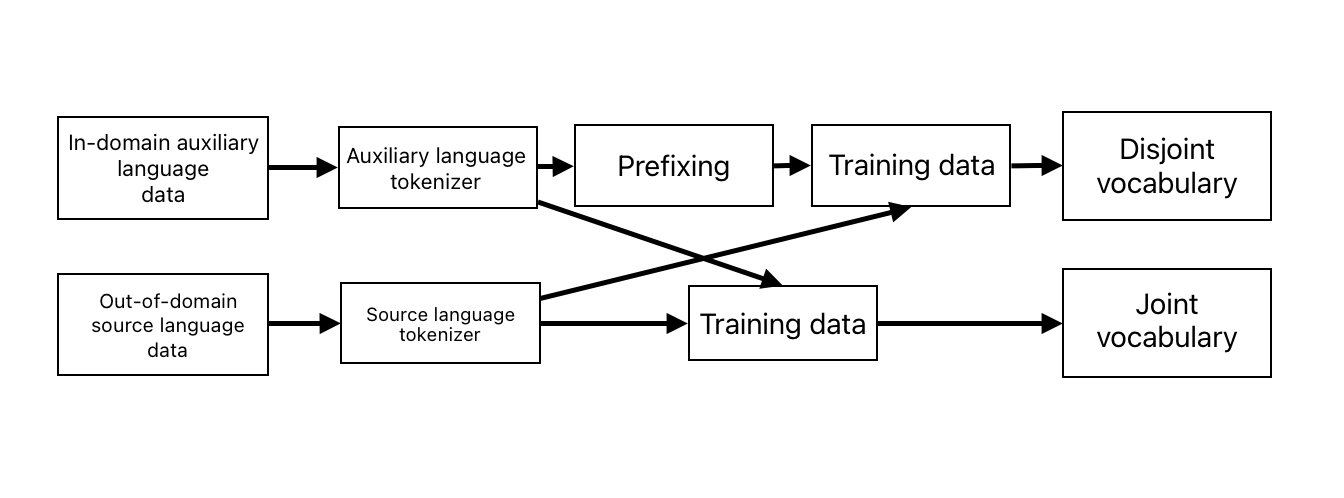}
  \caption{Illustration of our pipeline for extracting joint and disjoint vocabularies for training MNMT models.}
  \label{fig:pipeline}
\end{figure*}

\section{Methodology}
\label{sec:methods}

In this study, our aim is to find out how much knowledge transfer is due to vocabulary overlap by experimenting with auxiliary languages related and unrelated to the source language. Therefore, we need a comparable pipeline illustrated in Figure \ref{fig:pipeline} to experiment with joint and disjoint vocabularies. In this section, we describe our method for extracting vocabularies to train MNMT models.

\subsection{Out-of-Domain Setup}
To emphasize knowledge transfer in our experiments, we use an out-of-domain task and introduce in-domain data to our models only through auxiliary languages. For our training data, we used two different domains for the main and auxiliary language pairs. We describe the data used in our experiments in more detail in Section~\ref{sec:data}. 

\subsection{Tokenization}
\label{sec:tokenization}
Joint vocabularies are most often trained using a single multilingual tokenizer ~\cite{wu-monz-2023-beyond,imamura-utiyama-2024-empirical,kallini-etal-2025-false}. However, as we need to experiment with joint and disjoint vocabularies with a comparable design, we opt for language-specific tokenizer models that we applied separately to the respective data subsets.

To tokenize our data, we use SentencePiece ~\cite{kudo-richardson-2018-sentencepiece} with BPE and Byte-Fallback option, as suggested by Imamura and Utiyama~\shortcite{imamura-utiyama-2024-empirical}. We set character coverage to 1.0 and otherwise use the default settings. We use a vocabulary size of 32K for each language and for training the tokenizers, we use 1M lines of training data per language, except for the target language tokenizer, for which we use 3M lines in total, coming from the corresponding data to our source language and two auxiliary languages.

\subsection{Prefixing}
\label{sec:prefix}

In this study, we approach the issue of disjoint vocabulary by prefixing. We use a simple string substitution to add the prefix "AUX\_"  to every token in our auxiliary language data. Below, we present example sentences from the training data used in our experiments to demonstrate the effect. Tokens that would only be listed once in the vocabulary file (e.g. "in" and "ja") will be listed as two different vocabulary items as the result of prefixing.

\bigskip 

\noindent\textbf{German:}\\
▁Das ▁Parlament ▁hat \textbf{▁in} ▁diesem ▁Verfahren \textbf{▁ja} \\ ▁nur ▁die ▁Möglichkeit ▁der ▁Zustimmung ▁oder ▁der ▁Ablehnung .

\bigskip

\noindent\textbf{Swedish:}\\
▁Så ▁där ▁ja , ▁in ▁här .\\
SV\_▁Så  SV\_▁där \textbf{SV\_▁ja} SV\_, \textbf{SV\_▁in} SV\_▁här SV\_.

\bigskip

\noindent\textbf{Finnish:}\\
▁Muistakaa ▁äly , \textbf{▁in} to \textbf{▁ja} ▁its el uotta mus \\ FI\_▁Muistakaa FI\_▁äly FI\_, \textbf{FI\_▁in} FI\_to\\
\textbf{FI\_▁ja} FI\_▁its FI\_el FI\_uotta FI\_mus FI\_.

\bigskip

\subsection{Vocabulary Extraction}
\label{sec:vocab extraction}
Typically, vocabularies for MNMT can be taken directly from the tokenizer. However, since we use language-specific tokenizers, it is important to extract the vocabularies from tokenized corpora to properly contrast joint and disjoint vocabularies. After tokenization and prefixing, we merge the source and auxiliary language data to two different training data files per each language pair: one with and one without auxiliary prefixes. 

We then extract corresponding vocabulary files with the  marian-vocab tool \cite{mariannmt}. For target side, we follow the same procedure excluding the prefixing and for a bilingual baseline we exclude the auxiliary languge data. This method alters the size of the vocabularies to be slightly different from what is set when training the tokenizers. This is further discussed in Section \ref{sec:vocab size}. 

\section{Experiments}
\label{sec:experiments}

In this section, we describe our MNMT experiments using the pipeline presented in Section \ref{sec:methods}. We experimented with two auxiliary languages, Swedish and Finnish for translating from German to English. 

\subsection{Data} 
\label{sec:data}

For our multilingual models, we took half of the training data from the Europarl German--English dataset\footnote{\url{https://opus.nlpl.eu/Europarl/corpus/version/Europarl}}~\cite{TIEDEMANN12.463} and the other half from the auxiliary language OpenSubtitles2024 dataset\footnote{\url{http://www.opensubtitles.org/}\\
\url{https://opus.nlpl.eu/OpenSubtitles/corpus/version/OpenSubtitles}}~\cite{lison-tiedemann-2016-opensubtitles2016}, either Swedish--English or Finnish--English. 

For validation data in our models, we used 2000 disjoint lines from OpenSubtitles2024 dataset ~\cite{lison-tiedemann-2016-opensubtitles2016}. 
For the baseline we only used German-English data but for the multilingual models we used 1000 lines from the German–English dataset and 1000 lines from the auxiliary language datasets. All of the data for this study is downloaded from OPUS ~\cite{TIEDEMANN12.463} using OpusTools ~\cite{aulamo-etal-2020-opustools}. 

\subsection{Model Configurations}
\label{sec:config}
To carry out our experiments, we trained multiple translation models with MarianNMT~\cite{mariannmt}. All of our models are transformers and we used the Transformer-base architecture as described in Vaswani et al.~\shortcite{10.5555/3295222.3295349}. 

For training our models, we applied a configuration suggested in the GitHub of the OpusPocus -project \footnote{\url{https://github.com/hplt-project/OpusPocus/blob/main/ config/marian.train.teacher.base.yml}}. 
Our batch size was set to 4096 and the learning rate was 0.0003. We used the adam optimizer, with optimizer-delay of 2 and optimizer parameters 0.9, 0.98 and 1e-09. The number of training steps in our experiments varied between 75 000 and 175 000. We used the option ``shuffle: data’’ to shuffle our data before batching and we did not use tied embeddings to keep source and target embeddings independent.

In our experiments, the models were trained until convergence using early-stopping criteria on validation data. Validation was done at regular intervals (every 5000 steps) and we stopped the training procedures after 5 consecutive validation rounds without improvements in ChrF scores. The exact configuration file used in our experiments can be found in Appendix \ref{sec:appendix A}. The training was run on a single AMD MI250x GPU and each run took 10 hours on average.

\begin{table}
  \centering
  \begin{tabular}{llll}
    \hline
    \textbf{Model} & \textbf{Languages} & \textbf{Data} & \textbf{Vocabulary}\\
    \hline
    baseline   & de-en    &1M \\ 
    desv$\cup$1M  &   de+sv-en  & 1M   & joint\\
    defi$\cup$1M &   de+fi-en  & 1M &joint\\
    desv$\uplus$1M & de+sv-en  & 1M & disjoint \\
    defi$\uplus$1M & de+fi-en  & 1M & disjoint \\
    desv$\cup$ & de+sv-en  & 2M & joint \\
    defi$\cup$ & de+fi-en  & 2M   &joint\\
    desv$\uplus$   & de+sv-en & 2M   &disjoint\\
    defi$\uplus$ & de+fi-en   & 2M&disjoint\\
    \hline
    desv$\uplus$≠   & de+sv-en & 2M   &disjoint\\
    defi$\uplus$≠ & de+fi-en   & 2M&disjoint\\
    \hline
  \end{tabular}
  \caption{Amount of training data, languages and vocabulary settings of our experiments. Data refers to amount of training data in lines. $\cup$ refers to joint vocabulary and $\uplus$ refers to disjoint vocabulary. ≠ refers to unequal vocabulary size. For a complimentary experiment we manipulated the auxiliary 
  vocabulary sizes. }
  \label{tab:models}
\end{table}

\begin{table*}
  \centering
  \begin{tabular}{lllll}
    \hline
    \textbf{Model} & \textbf{Full size} & \textbf{Aux size} & \textbf{Overlap (tokens)} & \textbf{Overlap (percentage)}\\
    \hline
    baseline  &  31 421 &  &   &  \\ 
    desv$\cup$, desv$\cup$1M & 58 918  &  31 383 & 3 886 & 6.6\% \\
    defi$\cup$, defi$\cup$1M  & 60 577 & 31 671 & 2 515 & 4.2\% \\
    desv$\uplus$, desv$\uplus$1M  & 62 802 & 31 383 & 2 & 0.003\% \\
    defi$\uplus$, defi$\uplus$1M & 63 090 & 31 671 & 2 & 0.003\% \\
    \hline
    desv$\uplus$≠   & 58 427 & 27 008 & 2 & 0.003\% \\
    defi$\uplus$≠   &  60 271 & 28 852 & 2 & 0.003\% \\
    \hline
  \end{tabular}
  \caption{\label{Vocabulary size}
    Vocabulary sizes and amount of vocabulary overlap in our experiments. The models with 1M lines of training data use the same vocabulary files as the corresponding models with 2M lines of training data. Aux size refers to auxiliary language vocabulary size.
  }
\end{table*}

\subsection{Models}

Our primary experiments include eight different multilingual models and a bilingual German-English baseline. We first trained multilingual models with 2M lines of training data. As we were, however, restricted to train the baseline with only 1M lines due to the size of the Europarl dataset, we also trained multilingual models with only 1M lines of training data, divided as 500K lines from auxiliary language OpenSubtitles2024 datasets and 500K lines from German Europarl dataset, with their corresponding English data. For both auxiliary languages, we experimented with joint and disjoint vocabularies with both amounts of training data. All the different models are described in Table \ref{tab:models}. It is noteworthy, that our multilingual vocabulary files were extracted from the 2M lines of training data but were reused for training the otherwise corresponding models with 1M lines of data. 

\subsection{Vocabularies} 
\label{sec:vocab size}
As explained in Sections \ref{sec:tokenization} and \ref{sec:vocab extraction} we set our vocabulary size within SentencePiece to 32K for all of our languages, but derived the vocabulary files from our tokenized training data ending up with somewhat smaller vocabulary sizes.

For calculating the amount of vocabulary overlap in our experiments, we extracted vocabularies from the auxiliary language data. In Table \ref{Vocabulary size}, we present the sizes of our marian-vocab files and vocabulary overlaps in tokens and percentages.

The amount of vocabulary overlap in each model is the amount of overlapping tokens between the German vocabulary and the auxiliary language vocabulary. A formula for the amount of overlapping tokens, that is $|O|$, can be derived from the vocabulary size formula presented in section \ref{sec:background of vocab overlap} as follows: $|O| = |V^{de}|+|V^{aux}| - |V^{joint}|$. 

The models with disjoint vocabularies all seem to have two tokens overlapping despite the addition of the prefixes. These two tokens are an unknown token <unk> and a sentence boundary token </s>. When applying a joint vocabulary in our models, it seems that Swedish and German have a bigger overlap than Finnish and German. This is expected, as they are related and share more lexical items. 

\begin{figure}[t]
  \includegraphics[width=\columnwidth]{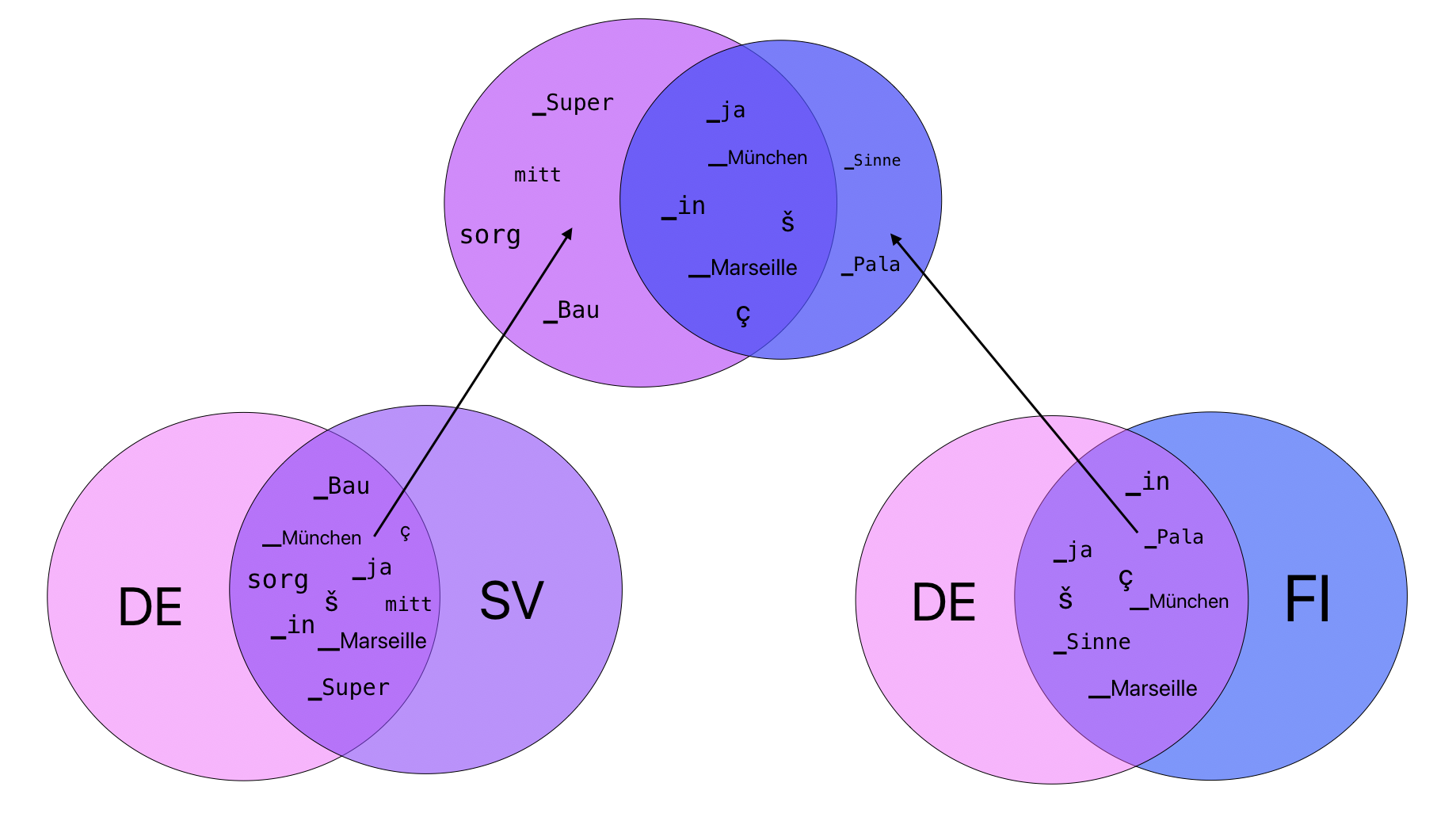}
  \caption{A figure illustrating the overlap of the vocabulary overlaps of German and Swedish and German and Finnish}
  \label{fig:overlapofoverlaps}
\end{figure}

To find out how much of the vocabulary overlap for German and Swedish and German and
Finnish is unique to those language pairs, we also looked into the overlap of the overlaps, illustrated in Figure \ref{fig:overlapofoverlaps}. The size of the overlap of the overlaps is 2072 tokens, and therefore the overlap of German and Swedish has 1814 tokens that are not in the overlap of German and Finnish and the overlap of German and Finnish has only
443 tokens that are not in the overlap of German and Swedish.

This means that for German and Finnish,
82\% of their overlap is also in the overlap of German and Swedish, while for German and
Swedish it is only 53\% of the overlap that is also in the overlap of German and Finnish. This shows that the overlap of German and Finnish adds only very little to the overlap of the
overlaps and that the amount of unique overlapping tokens, specific for German and Finnish, is
very small. The overlap of the overlaps seems to have a lot of proper nouns and tokens of length of only one or two characters. However, we leave further analysis of the overlapping tokens to future work.

\subsection{Complementary Vocabulary Size Experiments}
\label{sec:complementary}
We ran complementary experiments in which we manipulated the vocabulary sizes of our auxiliary language tokenizers. More precisely, our goal was to match the vocabulary sizes of the joint vocabularies even while having a disjoint vocabulary. We approached this by lowering the vocabulary size of the tokenizers for the auxiliary languages and otherwise following the same pipeline as we did with our primary disjoint vocabulary experiments presented in Section \ref{sec:methods} . We trained new tokenizers for the auxiliary languages with a vocabulary size equal to the difference between the size of the corresponding joint vocabulary and the size of the German vocabulary. More formally, given $|V^{joint}|$ is the size of the joint vocabulary $V^{joint}$, we trained the new auxiliary language tokenizers with the vocabulary size $|V^{aux}| =  |V^{joint}| - |V^{de}|$. 

As the decrease of the vocabulary size from what is set within SentencePiece to what marian-vocab produces persists, the vocabulary sizes of our joint vocabulary experiments and the vocabulary sizes we ended up with these complementary experiments do not add up perfectly, as shown in Table \ref{Vocabulary size}. 

\subsection{Evaluation}
\label{sec:evaluation}
We tested our models with the OpenSubtitles2024 German-English testset\footnote{\url{https://github.com/Helsinki-NLP/OpenSubtitles-devtest}}~\cite{OpenSubs2024:LREC2026}, which is a dedicated set of held-out data for testing translation in the subtitle domain, and evaluated the translations with sacreBLEU ~\cite{post-2018-call} using both BLEU ~\cite{papineni-etal-2002-bleu} and ChrF ~\cite{popovic-2015-chrf}. Although MT evaluation scores have been suggested to be not completely reliable metrics for knowledge transfer~\cite{stap-etal-2023-viewing}, we can see clear trends using BLEU and ChrF scores, which we describe in detail in the following section.

\section{Results}
\label{sec:results}
In this section, we present our results. All of our primary experiments were run six times, and we present the mean value of our BLEU and ChrF scores and the standard deviations.

\subsection{Improvement with an Auxiliary Language}

\begin{figure}[t]
\includegraphics[width=\columnwidth]{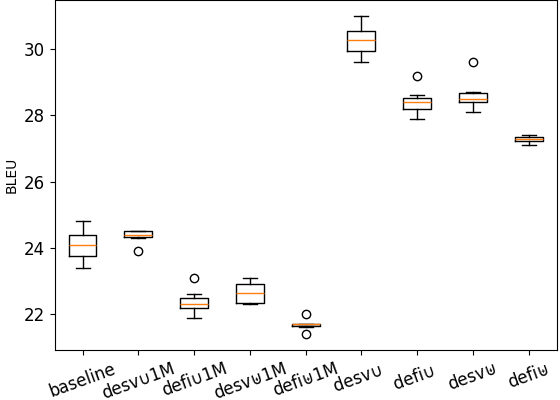}
  \caption{BLEU scores and standard deviations tested on OpenSubtitles2024 de-en testset of our primary results. A figure of ChrF scores illustrating the same trend can be found in the appendix \ref{sec:appendix B}.}
  \label{fig:bleu}
\end{figure}

As presented in Table \ref{tab:scores} and Figure \ref{fig:bleu}, we gained a remarkable improvement of scores by introducing the 1M lines of in-domain auxiliary language data to our models. Both Swedish and Finnish data increased the scores, as the average increase with an auxiliary language in a joint vocabulary setting is 5.3 in BLEU and 4.3 in ChrF. Swedish is, however, outperforming Finnish with almost 2 points in both metrics. The increase in using Swedish with a joint vocabulary compared to the baseline is 6.2 in BLEU and 5.2 in ChrF, while Finnish only increased by 4.3 in BLEU and 3.4 in ChrF.

The experiments with only 1M lines of multilingual training data give rather different results as the scores are mostly not even reaching the baseline. This is expected, as the models have only half the amount of German training data compared to their counterparts with 2M lines of training data. Measured in ChrF, both models have decreasing scores compared to baseline. 

As illustrated in Figure \ref{fig:bleu}, measured in BLEU here, combining German with Swedish outperforms the baseline. Based on this, it seems that in a scenario of related languages and a joint vocabulary, domain could be even more important than language, as the Swedish in-domain data seems to be able to outperform the German out-of-domain data. As we used the same vocabulary files regardless of the amount of training data, we cannot, however, argue to what extent the positive effect of the Swedish data is due to either vocabulary overlap or the vocabulary being trained with more data than the model itself.

In addition, the results are not enough to claim that the Finnish data would cause interference despite the decrease in scores, as the models trained with 2M lines of data outperform the baseline. The Finnish data can be of use, but it cannot replace the German data in the way Swedish data possibly can.

\subsection{Drop of Performance with a Disjoint Vocabulary} 

The use of a disjoint vocabulary decreased the scores compared to a joint vocabulary, but the models still outperform our baseline. As figure \ref{fig:bleu} illustrates, it seems that in our experiments vocabulary overlap was more beneficial for Swedish and German than it was for Finnish and German. Measured in BLEU, models with a disjoint vocabulary performed 1.4 points worse on average compared to their counterparts with a joint vocabulary. The difference is 1.7 points for German and Swedish and 1.1 points for German and Finnish. Measured in ChrF, the decrease is, however, only 0.9 on average. For German and Swedish, a disjoint vocabulary decreased the scores by 1.2 ChrF points, and for German and Finnish the difference was even smaller, only 0.6 points. As
the scores of the models trained with only 1M lines also show that the impact of joint vocabulary
is more significant for Swedish and German, there seems to be a pattern between the results and
the sizes of the vocabulary overlaps. Table \ref{Vocabulary size}, shows that the overlap is 6.6\% of the whole vocabulary with Swedish and German and 4.2\% for Finnish and German. 

However, we cannot argue that the difference in the size would be the only explanation behind this, as the semantic quality of the overlapping tokens might be affecting the results. Nevertheless, this finding supports the overall result of the models trained with Swedish and German performing better. This could be due to a cumulative effect stemming from non-lexical transfer and both the extent and semantical properties of the vocabulary overlap. 

As illustrated in Figure \ref{fig:bleu}, within the models trained with 2M lines of data, regardless of the vocabulary setting, the models trained with German and Swedish outperform the models trained with German and Finnish. Our results suggest that Swedish can bring more benefit even with a disjoint vocabulary than Finnish can with a joint one.

It is therefore evident that knowledge transfer is happening even with a disjoint vocabulary in our experiments, and other kind of similarity between languages must be beneficial in addition to lexical sharing. The hidden layers of a model are capable of transferring knowledge even when
vocabulary overlap is disabled, but based on our results this phenomenon is, however, not limited to only related languages. Regardless of the choice of auxiliary language, the disjoint vocabulary models trained with 2M lines of data are performing better than the baseline. This means that the in-domain knowledge of the auxiliary language data is transferred in them, although not quite to the same extent as with a joint vocabulary. 

\begin{table}
  \centering
  \begin{tabular}{lll}
    \hline
    \textbf{Model}           & \textbf{BLEU} & \textbf{ChrF} \\
    \hline
    baseline   & $24.1^{\pm 0.53}$ & $49.5^{\pm 0.31 }$ \\
    desv$\cup$1M &   $24.4^{\pm0.23}$ &$44.7^{\pm0.14}$\\
    defi$\cup$1M &   $22.4^{\pm0.42}$ & $42.9^{\pm0.37}$\\
    desv$\uplus$1M  &$22.7^{\pm0.36}$ & $43.1^{\pm0.34}$ \\
    defi$\uplus$1M & $21.7^{\pm0.19}$ &$42.1^{\pm0.2}$\\
    desv$\cup$ & $30.3^{\pm0.53}$ & $51.1^{\pm0.37}$ \\
    defi$\cup$  & $28.4^{\pm0.45}$ &$49.3^{\pm0.33}$\\
    desv$\uplus$  &$28.6^{\pm0.52}$ & $49.9^{\pm0.3}$ \\
    defi$\uplus$ & $27.3^{\pm0.12}$ &$48.7^{\pm0.4}$\\
    \hline
    desv$\uplus$≠ & 28.8 & 50  \\
    defi$\uplus$≠ & 27.3  & 48.9  \\
    \hline
  \end{tabular}
  \caption{BLEU and ChrF scores tested on OpenSubtitles2024 de-en testset. Mean and standard deviation are estimated over six runs. }
  \label{tab:scores}
\end{table}

\subsection{Results of Complementary Experiments}

As explained in section \ref{sec:complementary} we conducted complementary experiments in which we manipulated the vocabulary size in the auxiliary language to mitigate the potential impact of vocabulary size differences. 
 
Manipulation of the auxiliary vocabulary sizes did not seem to significantly alter the scores compared to disjoint vocabulary experiments, as shown in Table \ref{tab:scores}.\footnote{Note that the results are based on a single run in this case.} Interestingly, the scores even show slight improvements, which indicates that the loss in auxiliary language vocabulary coverage was not harmful for the model's performance. 

\subsection{Qualitative Analysis}
\label{sec:analysis}
To find concrete examples of knowledge transfer succeeding and failing in our multilingual experiments, we retrieved sentence pairs where models using a joint vocabulary performed 50 ChrF points better than models using a disjoint vocabulary. We yielded 90 pairs for German and Swedish and 100 for German and Finnish, some of which occurred in both. The examples presented in Table \ref{tab:examples} are unique to the language pair, but the observations that can be made based on them are rather similar regardless of the languages. The examples show that imperative mood with pronouns in accusative and dative case seem to cause confusion for the models using a disjoint vocabulary. This demonstrates that the colloquial style of movie subtitles is difficult to translate with an out-of-domain system, but that the use of a joint vocabulary seems to reduce such errors, which supports our finding of vocabulary overlap being beneficial for knowledge transfer. However, translation errors that arise from the use of a disjoint vocabulary deserve further investigation in future work.

\begin{table}[t]
  \centering
  \begin{tabular}{lll}
    \hline
     \textbf{Model }& \textbf{Translation} & \textbf{ChrF} \\
    \hline
    Source & "Lass mich dich überraschen." & \\
    Ref & "Let me surprise you." & \\
    desv$\cup$ & "Let me surprise you." & 100 \\
    desv$\uplus$ & 'I am surprised at you.' & 43.1 \\
    \hline
    Source & Gib mir einen Kuss. & \\
    Ref & Give me a kiss. & \\
    desv$\cup$ & Give me a kiss. & 100 \\
    desv$\uplus$ & I should like to make a point. & 12.9 \\
    \hline
    Source & -Ok, essen wir  & \\
    Ref & -Ok, let's eat.  & \\
    desv$\cup$ & - Oh, let's eat. & 74.6 \\
    desv$\uplus$ & -Oh, we eat. & 21.6 \\
    \hline
    Source & Halt dich an ihr fest! & \\
    Ref & Hold on to her! & \\
    defi$\cup$ & Hold on to her! & 100 \\
    defi$\uplus$ & Do not hold on to it! & 44.7\\
    \hline
    Source & Folge mir! & \\
    Ref & Follow me! & \\
    defi$\cup$ & Follow me!  & 100 \\
    defi$\uplus$ & My next point is this. & 3.1\\
    \hline
    Source & -Vorsicht, Eva. & \\
    Ref & -Be careful, Eva.  & \\
    defi$\cup$ & -Be careful, Eve.  & 84.8 \\
    defi$\uplus$ & - Precautionary, Eva. & 29.5\\
    \hline
  \end{tabular}
  \caption{Example translations with joint and disjoint vocabularies of sentences from OpenSubtitles2024 testset \cite{OpenSubs2024:LREC2026}.}
  \label{tab:examples}
\end{table}

\section{Conclusions}

In this study, we experimented with two different auxiliary languages, one related and one unrelated to the source language, and joint and disjoint vocabularies in out-of-domain machine translation. Our motivation was to find out how much knowledge transfer is due to vocabulary overlap. Based on our results, it seems that for both related and unrelated languages vocabulary overlap is beneficial but knowledge transfer persists even without it. Our main finding is, however, that both the lexical transfer from the vocabulary overlap and the transfer persisting even in the case of a disjoint vocabulary are more significant when the auxiliary language is related to the source language. Therefore, we conclude that domain-match and language relatedness is more important to knowledge transfer than a joint vocabulary. 

\section{Future Work}

As our experiments concentrate on measuring the improvement in scores and the size of the vocabulary overlaps, 
future work should expand this to investigate the quality of the phenomena, as well as extend our experiments to new language pairs. First, the semantic quality of the overlap and its significance for the transfer should be studied. Second, the mechanisms of non-lexical transfer should be unraveled to gain knowledge on what kind of similarity between the languages is most important for knowledge transfer to succeed.

\section*{Limitations}

The results and conclusions presented in this paper are limited to only two language pairs as source and auxiliary languages in only one translation direction, with limited data and vocabulary sizes. More languages and data would be needed to make our conclusions generalizable with more reliable patterns.

The qualitative analysis presented in Section \ref{sec:analysis} is only limited to subsets of the translated testset. Due to ChrF as the metric of evaluation being very sensitive to character level alterations the retrieved subsets include rather short sentence pairs, where one sentence most often matches the reference exactly and the other one does not. More diverse methods of analysis should be applied for more thorough understanding of errors stemming from the use of a disjoint vocabulary.

\bibliography{custom.bib}
\bibliographystyle{eamt26}

\appendix

\newpage
\section{MarianNMT configuration}
\label{sec:appendix A}

\begin{lstlisting}[frame=single]
train-sets:
  - traindata.deaux
  - traindata.deaux.en
vocabs:
  - deaux.yml
  - en.yml
tied-embeddings-all: false
workspace: 5000
optimizer-delay: 2
keep-best: true
overwrite: true
#no-restore-corpus: true
sigterm: save-and-exit
# Model
type: transformer
task: transformer-base

# Training
disp-freq: 1000u
save-freq: 5000u
max-length: 250
max-length-crop: false
shuffle: data
#sharding: global
#sync-freq: 200u
#cpu-threads: 0
mini-batch: 4096
mini-batch-words: 0
mini-batch-fit: true
mini-batch-fit-step: 10
sync-sgd: true
check-gradient-nan: true

# Validation
valid-sets:
  - validdata.de
  - validdata.de.en
valid-freq: 5000u
valid-metrics:
  - Chrf
  - bleu
  - ce-mean-words
valid-reset-stalled: true
early-stopping: 5
early-stopping-on: first
beam-size: 4
normalize: 1
valid-mini-batch: 64
valid-max-length: 1000

\end{lstlisting}

\newpage
\section{Results in ChrF}
\label{sec:appendix B}

\begin{figure}[h]
  \includegraphics[width=\columnwidth]{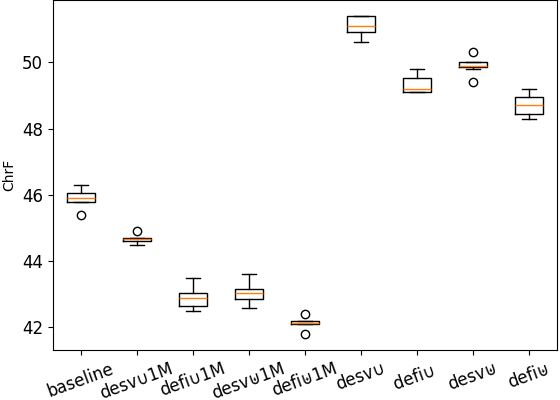}
  \caption{ChrF scores tested on the OpenSubtitles2024 de-en testset. In contrast to the BLEU scores in Figure \ref{fig:bleu}, the model trained with 1M lines of German and Swedish data did not outperform baseline.}
  \label{fig:chrf}
\end{figure}

\end{document}